\documentclass[runningheads]{llncs}
\usepackage[T1]{fontenc}
\usepackage{graphicx}
\usepackage[table]{xcolor}
\usepackage{multirow}
\usepackage{float}
\usepackage{subfigure}
\usepackage{url}
\usepackage{comment}
\usepackage{amsfonts}
\usepackage{makecell}

\newcommand\blfootnote[1]{%
  \begingroup
  \renewcommand\thefootnote{}\footnote{#1}%
  \addtocounter{footnote}{-1}%
  \endgroup
}

\begin{document}
\title{Next day fire prediction via semantic segmentation}
%
%
\author{Konstantinos Alexis\inst{2,3}
\and
Stella Girtsou\inst{1,4} \and
Alexis Apostolakis\inst{1,4} \and
Giorgos Giannopoulos\inst{2} \and
Charalampos Kontoes\inst{1}}
\authorrunning{K. Alexis et al.}
%
\institute{National Observatory of Athens, Greece  \and
Athena Research Center, Greece  \and
National Kapodistrian University of Athens, Greece \and
National Technical University of Athens, Greece 
}
\maketitle              
\begin{abstract}
In this paper we present a deep learning pipeline for next day fire prediction. The next day fire prediction task consists in learning models that receive as input the available information 
for an area up until a certain day, in order to predict the occurrence 
of fire for the next day. 
Starting from our previous problem formulation as a binary classification task on instances (daily snapshots of each area) represented by tabular feature vectors, we reformulate the problem as a semantic segmentation task on images; there, each pixel corresponds to a daily snapshot of an area, while its channels represent the formerly tabular training features. We demonstrate that this problem formulation, built within a thorough pipeline achieves state of the art results.\blfootnote{Preprint. Accepted in MACLEAN@ECML/PKDD 2023.}


\keywords{Next day fire prediction  \and Deep Learning \and Semantic Segmentation.}
\end{abstract}
%
%
%

\section{Introduction}
\label{sec:intro}

Forest fires are events with potentially catastrophic effects on social, environmental and economic level. Implementing methods and tools that are able to predict the occurrence of fire is of critical importance for public authorities and first responders (i.e. fire service) to plan their resources and operations, either long or short term. Long term planning includes fuel treatments, creation of fuel breaks or barriers around critical or sensitive sites or resources, vegetation modification, supervision of building activities to ensure adherence to fire safety standards and regulations and minimizing the potential for fire hazards, etc.
Short term planning involves the distribution of fire services forces at optimal locations, according to the estimated fire risks, as well as the proper communication/alerting of the public \cite{EASTAUGH2014132}. In this paper, we handle the problem of next day fire risk prediction, which serves the purposes of short term planning, in contrast to fire susceptibility, which serves long term planning. Next day fire prediction consists in learning a model that can exploit available information (meteorology, topography, vegetation, etc.)
for an area up until a certain day, in order to predict the occurrence, or probability of occurrence of fire for the \textit{next day}. Instead, fire susceptibility relaxes the prediction to a coarser time window in the future, e.g. a month or year.

We emphasize that next day fire risk prediction comprises a much more challenging task, mainly due to the following two reasons: (a) \textit{Extreme class imbalance}. Viewing the problem as binary classification for predicting fire risk on a large territory (e.g. a whole country) on a daily level, expectedly gives very few fire instances and many orders of magnitude more no-fire instances. Indicatively, in our dataset for the whole country of Greece, 
the ratio of fire to no-fire classes is in the order of $~1:100000$. It is well evident that most ML/DL approaches face issues when handling such imbalance, since they deploy accuracy-targeting optimization objectives and accuracy is an improper measure in cases of imbalance. (b) \textit{Extreme data scale}. A dataset in the order of billion instances poses several limits in performing proper cross validation processes for model selection/tuning and assessment. High Performance Computing (HPC) is not a solution that can be widely adopted in operation level, thus custom training and validation schemes need to be designed to optimize model training and selection. 

The vast majority of existing works either handle the much more relaxed problem of fire susceptibility or report next day fire prediction evaluation results on balanced or slightly imbalanced test sets, rendering their findings non-generalizable in real world distributions, as discussed in Section \ref{sec:rel}.

In our previous work \cite{rs14051222} we have presented state of the art results by implementing a complete ML pipeline that handled the problem as a binary classification task on fire-driving induced feature vectors representing daily snapshots of grid cells-areas.
In the current work we take a different approach, by reformulating the initial \textit{binary classification on tabular data} task into a \textit{semantic segmentation on images} task. We deploy a U-net architecture and we frame it in a pipeline, which performs \textit{task specific} dataset preprocessing and selection; training/validation data splitting and augmentation; and model tuning/selection. To the best of our knowledge, this is the first work that properly adapts semantic segmentation methods on the problem of next day fire prediction. We note that, although there exist previous works deploying semantic segmentation on the general task of fire risk prediction/estimation, the novelty of our proposed method is that it properly 
integrates a semantic segmentation model on a task specific ML pipeline, including dataset selection, data augmentation and task specific measures for model selection. Our results clearly outperform the ones of \cite{BHOWMIK2023117908}, as well as demonstrate considerable improvements compared to our 
previous results of \cite{rs14051222}.




\section{Background and related work}
\label{sec:rel}

\subsection{Problem definition}
\label{sec:problem}

We consider a geographic grid of square cells covering a territory of interest.
Each instance corresponds to the daily snapshot of each grid cell $i$, and is represented by a set of characteristics (alt. fire driving factors, features) that are extracted for the specific area, for a specific day $k$.
Given a historical dataset annotated (labeled) with the existence or absence of fire, for each grid cell, for each day, each available historical instance carries a binary label $l_{ik}$ for the day $k$, denoting the existence (label:fire) or absence of fire (label:no-fire). It is important to point out that all the features of the instance $x_{ik}$, are available from the previous day $k-1$. 
Thus, our problem is formulated as a classification task; the goal is to learn, using historical data, a decision function $f_H(x_{ik}:\theta)$, comprising a set of hyperparameters $H$ to be properly selected and a set of parameters $\theta$ to be properly learned, that, given a new instance $x_{ik}$ accurately predicts label $l_{ik}$.

In our initial formulations, $x_{ik}$ comprised a feature vector consisting of either numerical or categorical, one hot encoded data. In the current work, for each day, we consider the whole territory as an image and the grid cells as its pixels. Each channel of the image corresponds to one of the features. Consequently, an image of the whole territory is produced per day, and each instance of each image represents an instance $x_{ik}$, for which a class needs to be predicted. Given this, to solve the task, function $f_H(x_{ik}:\theta)$ needs to be derived from the family of semantic segmentation models—convolutional neural network architectures designed to perform classification on the pixel level of an image.

\subsection{Related work}
\label{sec:relwork}

Research on fire risk modeling the last decade is shifting towards machine learning (ML) or Deep learning (DL), due to the advantage of those methods in identifying non-linear relationships between predictor variables as opposed to classical statistical methods. The physical/theoretical models, on the other hand, lag behind ML/DL in that they cannot incorporate all the different types of the predictor variables (e.g. meteorology, terrain, satellite indices) that are important for the prediction \cite{Jain2020},\cite{bui2017hybrid}.

The extreme class imbalance 
is handled in the vast majority of studies by severely undersampling the no-fire class. 
While this practice may be a good enough approximation for solving problems where we calculate the fire occurrence probability for longer periods (season, years), in the case of the next day's problem, this leads to reporting results that are not representative enough of the real world setting. 
To our knowledge, a CNN model for fire susceptibility mapping was first introduced in \cite{zhang_forest_2019}. The training dataset was created by undersampling the no-fire class for achieving a balanced class distribution, including the test set. The CNN outperformed the traditional ML models it was compared with and the fire prediction period was yearly. A susceptibility map was presented for the whole dataset, however the performance results (Sensitivity 0.92, Specificity 0.83) referred to the validation dataset only. \cite{Huot2020segmentation} presented a semantic segmentation method for three problem variations predicting fire masks (i) daily, (ii) 7 days, and (iii) 7 days predicted from time series organized input. The architectures employed consisted of an Autoencoder, a U-Net and Autoencoder-LSTM plus Autoencoder-U-Net for the time series dataset. The test dataset comprised of tiles that were undersampled on no-fire class to reach a 1:2 fire/no-fire ratio. Of course, pixel wise, a far higher imbalance for the semantic segmentation task remained. The results for the best model variation gave AUC 0.83 and sensitivity 0.12, with the latter being a rather low score for real world utilization. \cite{Santopaolo2021} used a CNN encoder-decoder architecture for applying semantic segmentation to produce a fire prediction mask of an 8-days period. The model was trained using a custom weighted MSE loss function, giving more weight to the fire areas for addressing the imbalance problem. For the evaluation, a methodology was proposed that considered a fire prediction as true positive if it lies within a buffer zone (of 30px) around an actual fire pixel. No metrics were finally presented, while some selected sample masks were provided for visualizing the prediction in relation to the ground truth. \cite{prapas_deep_2021} handled the next day prediction task using CNN, LSTM and ConvLSTM architectures. The CNN tiles (for CNN and ConvLSTM) were formed around the predicted pixel. The metrics of the best models Sensitivity 0.762 (LSTM) and AUC 0.926 (ConvLSTM) were reported on an undersampled (1:3 fire/no-fire) test dataset. 



Regarding works that properly handle next day fire prediction, to the best of our knowledge, our works in \cite{9554301},\cite{rs14051222} handle the problem in its realistic basis. In particular, in these works we learned binary classifiers on an extended set of fire driving factors-features and assessed them strictly on holdout test sets that maintained the extreme imbalance of real world data. We built these models within a complete ML pipeline that properly avoids pitfalls of information leakage due to spatio-temporal correlations inherent within the data, as well as tries to compensate for the data imbalance during the cross validation process, by applying custom measures for model validation and selection. Apart from the above, \cite{BHOWMIK2023117908} appears to properly formalize a realistic next day prediction problem, although details on the distribution of the dataset are not clear. The authors implement DL architectures including U-net CNNs and U-net LSTMs and assess their models in holdout test sets, measuring the accuracy of prediction on image pixels representing $2.5X2.5$km areas. However, the sensitivity (recall of fire class reported) is quite low: $60\%$ and $42\%$ for large and small fires respectively, while specificity (recall of no-fire class) results are not reported.


\section{Methodology}
\label{sec:method}

\begin{figure}[H]
\centering
\includegraphics[width=12cm]{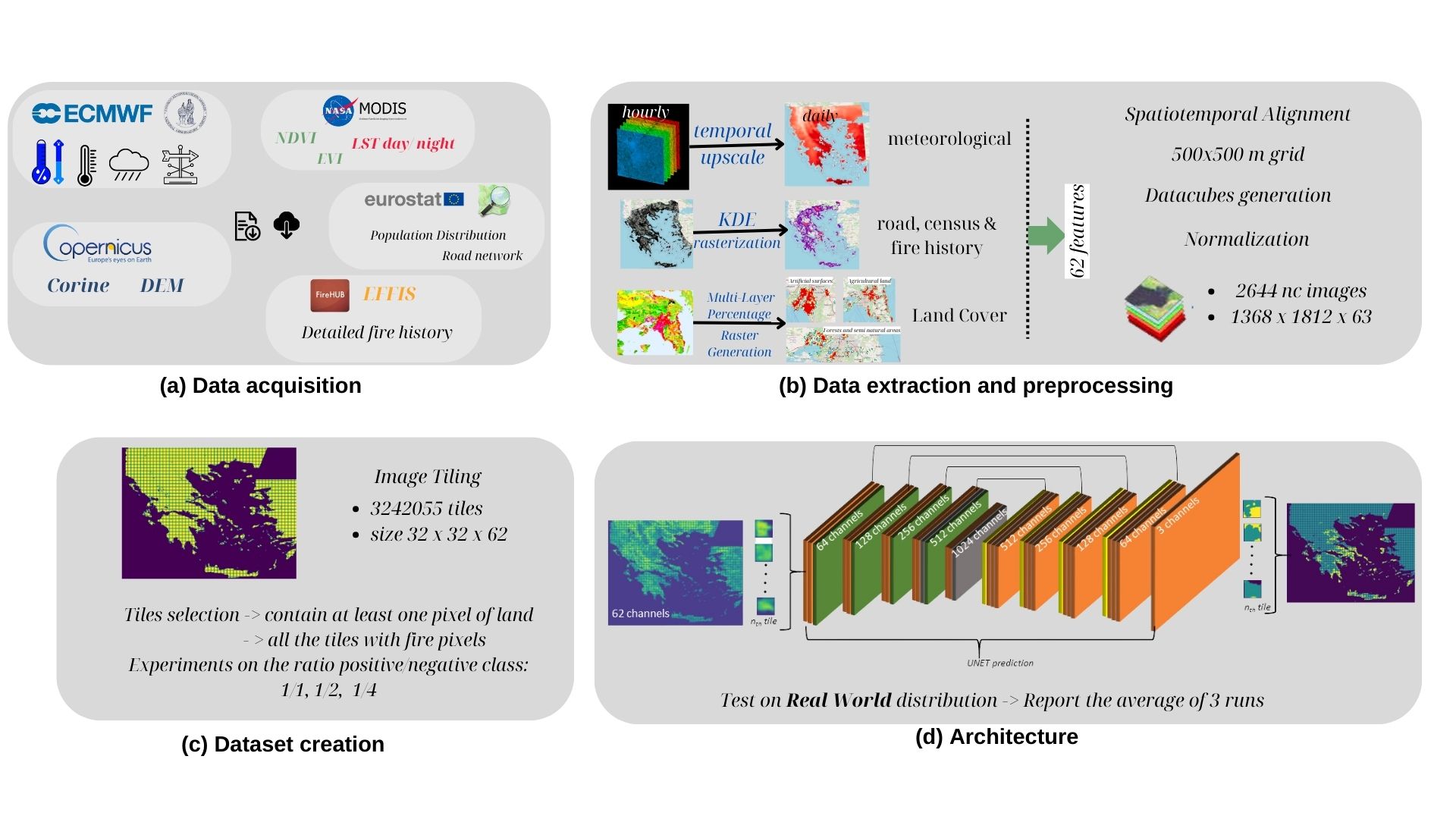}
\caption{The implemented semantic segmentation pipeline.} 
\label{fig1}
\end{figure}

An overview of the proposed methodology is shown in Figure~\ref{fig1}. We aim to learn semantic segmentation models that are able to effectively identify contextualized feature patterns corresponding to high fire risk. Our method starts by performing feature extraction from fire driving data sources to obtain proper feature representations for the instances of the task (daily snapshots of grid cells). It then applies a data pre-processing pipeline in order to come up with a segmentation compatible dataset. Then, it instantiates a U-Net architecture, within a cross-validation procedure that selects an optimal (trained) model, according to a task specific measure. Finally, the selected method is assessed on a holdout dataset that maintains the initial class distribution at the examined territory. 


\subsection{Feature extraction}
\label{sec:feat}
A series of fire driving factors were considered, by fusing datasets from various 
sources, including FireHub, EFFIS, NASA, ERA5, Copernicus, Open Street Map (OSM) and Eurostat as explained in \cite{10.1007/978-3-030-67835-7_27}. Out of them, a set of $62$ training features were extracted. These are briefly enumerated in the list below, while detailed information can be found in \cite{10.1007/978-3-030-67835-7_27}, \cite{rs14051222}.

\begin{itemize}
    \item \textbf{Digital elevation level}. This includes elevation and three more derivative factors: aspect, slope, curvature provided by Copernicus
     \item \textbf{Land cover}. Land cover type; this is provided by Copernicus’ Corine Land Cover\footnote{https://land.copernicus.eu/pan-european/corine-land-cover}
     \item \textbf{EO data}. NDVI, EVI, LST Day and Night from MODIS sensor onboard AQUA and TERRA satellites provided by NASA
     \item \textbf{Historical Weather predictions}. This includes 2m temperature, 2m dew temperature, 10m u-component of wind, 10m v-component of wind and total precipitation from ERA-5 Land. 
     \item \textbf{Census data}. Gridded cencus data from Eurostat\footnote{\urlstyle{same}\url{https://ec.europa.eu/eurostat/web/gisco/geodata/reference-data/grids}} available for 2011 and 2018.
     \item \textbf{Road density layer}. This was calculated by applying Kernel density function on the road network provided by OSM
     \item \textbf{Spatially smoothed fire history}. The fire history from 1984 to 2009 was obtained from FireHUB BSM\footnote{\urlstyle{same}\url{http://ocean.space.noa.gr/diachronic_bsm/}} service and was smoothly dispersed by the application of a low-pass linear filter.
     \item \textbf{Cell coordinates}. The latitude and longitude values of the cells' centroids.
     \item \textbf{Month of the Year and Week day}.
\end{itemize}



\subsection{Dataset selection and augmentation}
\label{sec:dataselect}
As a first step, we transform our available grid-based data collection into a dataset that can be utilized in a semantic segmentation setting. 
Feature scaling is performed for numerical features via min-max normalization, and one-hot encoding is applied on categorical ones.

Next, for each day of the dataset, the whole examined territory is transformed into an image of dimensions $h*w*62$, with $h$ being the height of the image in pixels (500m wide), $w$ being its width and $62$ its channels corresponding to the features extracted, as described in Section \ref{sec:feat}. Another image of dimensions $h*w*1$, holds the labels for each pixel of the former image, comprising its \textit{mask}.

Then, from these daily image-mask pairs, we extract tiles of shape $32*32$. The pixels composing these tiles correspond to the initial grid cells. Out of these tiles, we maintain in our train-validation dataset all the tiles containing at least one pixel of class fire, which we term as \textit{fire-tiles}. We dispose all tiles that are covered exclusively by water, since these are useless in our setting: there is zero chance that a fire occurrence is identified in any pixel of these tiles. Additionally, for tiles containing land but no fire classes, referred to as \textit{no-fire-tiles}, we randomly sample a subset to include alongside the fire-tiles. The ratio of no-fire to fire tiles comprises a hyperparameter of our pipeline which we intend to experiment with more in future steps.

We note the sampling process is performed \textit{exclusively on the train-validation set}. The holdout sets are left as is with respect to tile sampling, maintaining thus their initial, real world distribution. 






\subsection{Semantic segmentation network training and validation}
\label{sec:semseg}

\subsubsection{Semantic segmentation architecture: U-Net}
\label{sec:unet}
In this work, we leverage the U-Net \cite{ronneberger2015u} architecture as our semantic segmentation model. U-Net is a convolutional neural network architecture widely employed in the field of image segmentation and other pixel-wise prediction tasks. The network consists of an encoder-decoder architecture, capturing contextual information through its encoder layers and then leveraging skip connections to maintain spatial details for precisely localize and segment objects in input images.

After assessing the performance of various U-Net configurations, we opted for a default instantiation with a symmetrical architecture. The model consists of four down-sampling blocks within the encoder, each composed of two 3x3 convolutional layers followed by ReLU activation, followed by a 2x2 max-pooling layer. This design progressively reduces spatial dimensions by a factor of 2 at each down-sampling block. The decoder encompasses four up-sampling blocks, each comprising a transposed convolutional layer with a stride of 2, followed by a 3x3 convolutional layer with ReLU activation. Concerning the channel configuration, we initially used the default setting of 64 features in the first encoder block. Subsequently, we also experimented with 128 initial features to accommodate the complexity of our 62-channel input images.

A direct gain we expect from switching to a semantic segmentation formulation is the inherently better handling of spatial correlations by the deployed convolutional models, compared to the previous formulation that treats each instance independently.


\subsubsection{Cross-validation and model selection}
\label{sec:training}
We implement a cross-validation process for training the U-Net architecture on a training subset and performing model selection, via early stopping, on a validation subset. The training dataset is split into $k$ folds and, at each of the $k$ training-validation iterations, one fold is used for early stopping of the training process that is performed on the rest folds. We use k-fold splitting, instead of a single train-validation split, to average the measured validation scores over the $k$ iterations, to account for the stochasticity of the learning process. We note that further examining, reporting and mitigating this matter, e.g. by fine-tuning the optimization parameters of the U-net to achieve more stable behavior/convergence through training epochs comprises part of our future work. 

Each time, data augmentation for minority class (fire) oversampling is performed exclusively on the train set. In particular, a buffer of adjacent pixels on every initial fire pixel is created and the class of all pixels contained in the buffer is transformed to fire. This follows the intuition of the concept of ``absence of fire'' \cite{rs14051222}, according to which, such close pixels to a fire pixel most probably had similar conditions to the fire pixel, that could allow fire occurrence also on them, but due to factors that cannot be captured by the considered feature set, fire eventually did not occur. We note that this buffering procedure is also a hyperparameter in our pipeline and one can select whether to apply it or not on the train/validation sets individually. Our future work involves further experimentation with these options.
To mitigate class imbalance in the model training, we apply cost-sensitive learning, by setting class weights equal to the inverse class frequencies, on pixel level, in the train set.
In our experiments, the learning rate is set to 0.001, utilizing the cross-entropy loss function as the criterion and minimizing it with the Adam optimizer.

Finally, model selection is performed by early stopping on the validation set. The early stopping condition examines whether a selected measure is not improved for a certain number of epochs (also a hyperparameter) and then selecting the lastly improved checkpoint of the model being learned. In all cases, we train the models for 45 epochs at most.
We adopt the $shybrid_l$ ($sh_l$) measure from \cite{rs14051222}, which allows us, by varying $l$ to directly perform model selection on a weighted average of sensitivity and specificity, choosing whether and to what extent we wish to favor sensitivity:

\begin{equation}
\label{custom-harm}
shybrid_l=l*sensitivity+specificity
\end{equation}




\section{Experimental evaluation}
\label{sec:exp}

\subsection{Evaluation setting}
\label{sec:setting}

Our evaluation dataset considers the whole territory of Greece, defining a grid of cells 500m wide. This results to a total of $\sim 889$K cells per day. We consider the four months of June-September, since this is the time period when the vast majority of fire takes place in Greece, for the years 2010-2020. This gives us a total dataset of $\sim 1.19$ billion instances, i.e. daily snapshots of a cell.
From this, we leverage two evaluation settings, i.e., keeping years 2010-2018 and 2010-2019 as train-validation datasets, and treating years 2019 and 2020 as the corresponding holdout test sets.
In a typical experiment within our setting, this results in approximately 3.5K and 155K image tile instances used for training and testing, respectively.
Statistics on the fire and no-fire classes on the selected dataset are provided in Table \ref{tab:dataset}.

\begin{table}[H]
\centering
\caption{
Distribution of fire and no-fire instances. ``M'' stands for Millions.\label{tab:dataset}}
\begin{tabular}{|c|c|c|c|c|c|c|c|c|c|c|c|c|}
\hline 
\multicolumn{1}{|l}{\textbf{Year}} & & \textbf{2010} & \textbf{2011} & \textbf{2012} & \textbf{2013} & \textbf{2014} & \textbf{2015} & \textbf{2016} & \textbf{2017} & \textbf{2018} & \textbf{2019} & \textbf{2020} \\
\hline
\multicolumn{1}{|l}{\textbf{Fire}} & & 697 & 2522 & 3122 & 1345 & 1015 & 846 & 2076 & 1663 & 646 & 682 & 828 \\
\hline
\multicolumn{1}{|l}{\textbf{No-Fire}} & & 108M & 108M & 108M & 108M & 108M & 108M & 108M & 108M & 108M & 108M & 108M \\
\hline
\end{tabular}
\end{table}

Deploying a next day fire prediction system in operational environment has particular requirements regarding the prediction effectiveness, which we have derived through our collaboration with the Greek Fire Service. The foremost one is that, the majority of fire events are predicted by the system; this translates to high \textit{sensitivity} (recall of the fire class), that is defined as the ratio of predicted fire instances to the number of actual, total fire instances. On the other hand, due to the limited resources (vehicles, human power), it is not affordable to obtain high fire risk predictions for a large percentage of the country, since it will be infeasible by the fire service to cover all high risk areas; this translates to high \textit{specificity} (recall of the no-fire class), that is defined as the ratio of predicted no-fire instances to the number of actual, total no-fire instances. 

The above requirements are in general contradicting.
The exact trade-off between these measures might vary according to the exact operational requirements and context, being affected for example by the extent of reach of different fire services, the differences in resources and strategies, etc. 
Indicatively, according to the feedback we have received from the Greek Fire Service, a balanced set of sensitivity/specificity values, e.g. $(80\%,80\%)$, is considered preferable than a less balanced on, e.g. $(90\%,70\%)$.
Consequently, for the assessment of such methods we firmly believe that the joint inspection of the two individual measures of sensitivity and specificity is required.
Thus, we choose not to report on a single measure, e.g. balanced accuracy \cite{5597285} and, on the contrary, show results that present different balances between the two individual measures.

\subsection{Results}
\label{sec:results}

We assess the performance of models variations that are either optimized for high sensitivity, or towards a balance among sensitivity and specificity. This is done in order to accommodate the various use-cases that the current method can implement.

\begin{table}[H]
\centering
\caption{
Validation results for the proposed method with different configurations. A configuration is defined by the Tile Ratio (\textbf{TR}), which specifies the ratio of \textit{no-fire-tiles} to \textit{fire-tiles} in the dataset; the use of Fire Buffer (\textbf{FB}) for augmenting fire instances; the number of Initial Features (\textbf{IF}) in the U-Net architecture; and the Early Stopping (\textbf{ES}) metric, which determines the monitored metric for early stopping. Results are averaged over three validation folds.
\label{tab:val_results}}
    \begin{tabular}{cccccccccccc}
    \hline \hline
    \multicolumn{4}{c}{\textbf{Config.}} & \multicolumn{4}{c}{\textbf{June-Sept. 2010-2018}} & \multicolumn{4}{c}{\textbf{June-Sept. 2010-2019}} \\
    \textbf{TR} & \textbf{FB} & \textbf{IF} & \textbf{ES} & \textbf{Sens.} & \textbf{Spec.} & \textbf{sh1} & \textbf{sh2} & \textbf{Sens.} & \textbf{Spec.} & \textbf{sh1} & \textbf{sh2}\\
    \hline
    1 & $\times$     & 64 & sh1  & 0.8379          & 0.7007          & 1.5386          & 2.3765 & 0.8564          & 0.6768          & 1.5332          & 2.3896          \\
    1 & $\times$     & 128 & sh1 & 0.7976          & 0.7379          & 1.5356          & 2.3332 & 0.7993          & 0.7288          & 1.5281          & 2.3275          \\
    1 & \checkmark & 64 & sh1  & 0.8450          & 0.6999          & 1.5449          & 2.3899 & 0.8344          & 0.7023          & 1.5366          & 2.3710          \\
    1 & \checkmark & 128 & sh1 & 0.8652          & 0.6856          & 1.5508          & 2.4160 & 0.8203          & 0.7153          & 1.5356          & 2.3559          \\
    4 & $\times$     & 64 & sh1  & 0.9033          & \textbf{0.7866} & 1.6899          & 2.5933 & 0.9090          & 0.7855          & 1.6945          & 2.6035          \\
    4 & $\times$     & 128 & sh1 & 0.9069          & 0.7746          & 1.6815          & 2.5884 & 0.8823          & \textbf{0.8060} & 1.6883          & 2.5706          \\
    4 & \checkmark & 64 & sh1  & 0.9114          & 0.7769          & 1.6883          & 2.5997 & 0.9196          & 0.7878          & \textbf{1.7074} & \textbf{2.6270}          \\
    4 & \checkmark & 128 & sh1 & \textbf{0.9171} & 0.7734          & \textbf{1.6905} & \textbf{2.6075}& \textbf{0.9208} & 0.7784          & 1.6992          & 2.6200 \\
    \hline
    1 & $\times$     & 64 & sh2  & 0.8980          & 0.6168          & 1.5148          & 2.4129 & 0.9393          & 0.5256          & 1.4649          & 2.4041          \\
    1 & $\times$     & 128 & sh2 & 0.8866          & 0.6234          & 1.5100          & 2.3966 & 0.8663          & 0.6610          & 1.5273          & 2.3936          \\
    1 & \checkmark & 64 & sh2  & 0.9375          & 0.5719          & 1.5093          & 2.4468 & 0.9106          & 0.6039          & 1.5145          & 2.4251          \\
    1 & \checkmark & 128 & sh2 & 0.9371          & 0.5456          & 1.4827          & 2.4198 & \textbf{0.9478} & 0.5354          & 1.4832          & 2.4309          \\
    4 & $\times$     & 64 & sh2  & 0.9396          & 0.7242          & 1.6638          & 2.6033 & 0.9308          & 0.7463          & 1.6772          & 2.6080          \\
    4 & $\times$     & 128 & sh2 & 0.9444          & \textbf{0.7316} & 1.6760          & 2.6204 & 0.9474          & 0.7210          & 1.6684          & 2.6157          \\
    4 & \checkmark & 64 & sh2  & 0.9566          & 0.7224          & 1.6791          & 2.6357 & 0.9441          & \textbf{0.7513} & \textbf{1.6954} & \textbf{2.6394}          \\
    4 & \checkmark & 128 & sh2 & \textbf{0.9567} & 0.7229          & \textbf{1.6796} & \textbf{2.6363} & 0.9445          & 0.7431          & 1.6876          & 2.6321  \\
    \hline
    \end{tabular}
\end{table}

As shown in Table~\ref{tab:val_results}, we subject the proposed method to a variety of configurations for both evaluation settings, prioritizing either \textit{sh1} or \textit{sh2} and accordingly striking a balance or favoring sensitivity over specificity. The results reveal that each configuration choice significantly impacts the method's performance. Notably, configurations involving a higher proportion of \textit{no-fire-tiles} (four times the number of \textit{fire-tiles}) and the application of the fire buffer augmentation are consistently involved in the best results when optimizing for either \textit{sh1} or \textit{sh2}. These findings demonstrate the effectiveness of these 
choices.

Based on the validation results, we identify the best performing instantiations of our method, and select them for evaluation on the holdout test sets.
We compare the current results with two of the best performing models that were fine-tuned for hyperparameterization in \cite{rs14051222}; the high sensitivity oriented (NNd-sh2), which is a model with a fully-connected architecture and dropout layers optimized for the \textit{sh2} metric and the balanced (NNd-auc), which is a similar model optimized for AUC.
As shown in Table~\ref{tab:test_results}, our current approach results in improved models, outperforming our previous method in the balanced scenario while remaining competitive in the sensitivity-favoring setting.

\begin{table}[H]
\centering
\caption{
Sensitivity and specificity of the proposed method, compared to \cite{rs14051222} on the 2019 and 2020 test sets. Rows 1 and 3 report on models selected on the validation set, using measures that slightly favor sensitivity against specificity, for our previous and current work respectively. Rows 2 and 4 report on models that seek a better balance between sensitivity and specificity. In both cases, our current method matches or even considerably improves upon our previous one.
\label{tab:test_results}}
\begin{tabular}{cccccc}
\hline \hline
\multirow{2}{*}{\textbf{\#}} & \multirow{2}{*}{\textbf{Algorithm/Model}} & \multicolumn{2}{c}{\textbf{June-September 2019}} & \multicolumn{2}{c}{\textbf{June-September 2020}} \\
{} & {} & \textbf{Sens.} & \textbf{Spec.} & \textbf{Sens.} & \textbf{Spec.} \\
\hline
1 & NNd-sh2 & 0.88 & 0.83 & 0.93 & 0.85 \\
2 & NNd-auc & 0.81 & 0.83 & 0.83 & 0.88 \\
3 & U-Net-sh2 (current) & 0.95 & 0.78 & 0.93 & 0.84 \\
4 & U-Net-sh1 (current) & 0.87 & 0.83 & 0.89 & 0.87 \\
\hline
\end{tabular}

\end{table}

\vspace{-2ex}

\begin{figure}[h]
    \centering
    \begin{tabular}{cc}
        \textbf{\scriptsize \quad \quad Ground Truth  \quad Model Prediction}&\textbf{\scriptsize \quad \quad Ground Truth \quad Model Prediction} \\
        \includegraphics[width=0.45\textwidth]{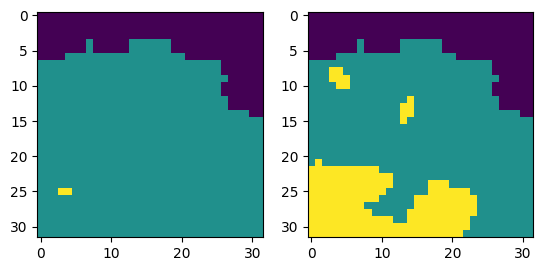} &
        \includegraphics[width=0.45\textwidth]{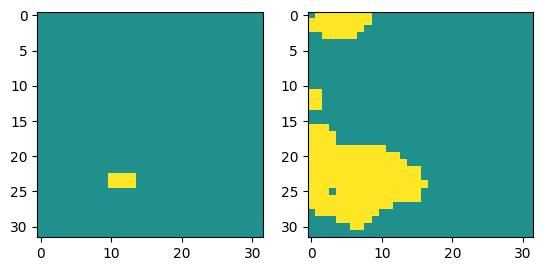} \\
        \includegraphics[width=0.45\textwidth]{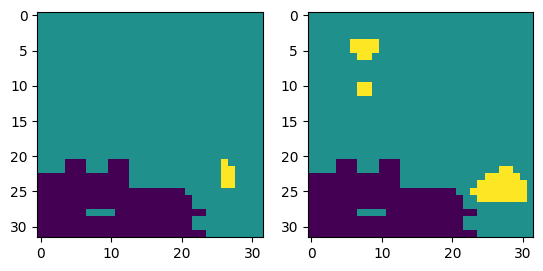} &
        \includegraphics[width=0.45\textwidth]{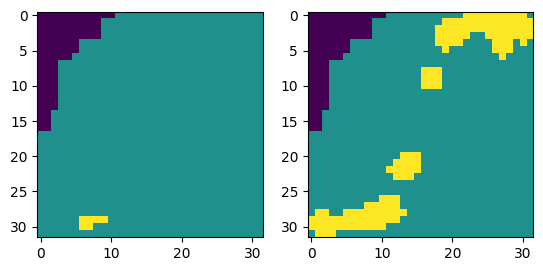} \\
    \end{tabular}
    
    \caption[short]{Predicted fire masks on test tiles. Images on the first and third columns correspond to the actual fire annotations, while on the second and fourth columns the corresponding predicted masks are shown. Images on the first row showcase predictions made by the U-Net-sh2 variation proposed in this work on random 2019 test tiles. In the second row, predictions made by the U-Net-sh1 on random 2020 test tiles are shown.}
\label{fig2}
\end{figure}

Figure~\ref{fig2} demonstrates for a sample of fire test tiles, their true label masks and the corresponding predicted ones made by our models. It is clear that the models are able to capture the fine-grained fire patterns adequately, even though the latter comprise only a small percentage of the input image tiles.

\vspace{-2ex}

\section{Conclusion}
\label{sec:concl}
In this paper, we proposed a semantic segmentation formulation for the next day fire prediction task. We transformed historical tabular data into image-mask pairs and trained a series of U-Net models for this segmentation task in a supervised way, taking into account the task specificities to perform appropriate dataset selection and augmentation and model selection. The proposed method improves upon the evaluation metrics, while also being faster to train and more robust compared to our previous best approach for the task. Our next steps involve performing a more thorough hyperparameter search and model tuning/selection process, maintaining the real world distribution of the test set also on the the validation set, so that we can directly perform model selection on the same distribution, as well as examining ensemble models with the traditional ML approach of our previous work \cite{rs14051222}.

%
%


\subsubsection{Acknowledgements}
This work has been supported by the national research project PREFERRED, which is co-funded by Greece and the European Union through the Regional Operational Programme of Attiki, under the call "Research and Innovation Synergies in the Region of Attica” (Project code: ATTP4-0340489).

%
%
%
%

\bibliographystyle{plain}
\bibliography{bibliography}

\begin{thebibliography}{10}

\bibitem{rs14051222}
Alexis Apostolakis, Stella Girtsou, Giorgos Giannopoulos, Nikolaos~S.
  Bartsotas, and Charalampos Kontoes.
\newblock Estimating next day's forest fire risk via a complete machine
  learning methodology.
\newblock {\em Remote Sensing}, 14(5), 2022.

\bibitem{10.1007/978-3-030-67835-7_27}
Alexis Apostolakis, Stella Girtsou, Charalampos Kontoes, Ioannis Papoutsis, and
  Michalis Tsoutsos.
\newblock Implementation of a random forest classifier to examine wildfire
  predictive modelling in greece using diachronically collected fire occurrence
  and fire mapping data.
\newblock In {\em MultiMedia Modeling}, pages 318--329, Cham, 2021. Springer
  International Publishing.

\bibitem{BHOWMIK2023117908}
Rohan~T. Bhowmik, Youn~Soo Jung, Juan~A. Aguilera, Mary Prunicki, and Kari
  Nadeau.
\newblock A multi-modal wildfire prediction and early-warning system based on a
  novel machine learning framework.
\newblock {\em Journal of Environmental Management}, 341:117908, 2023.

\bibitem{5597285}
Kay~Henning Brodersen, Cheng~Soon Ong, Klaas~Enno Stephan, and Joachim~M.
  Buhmann.
\newblock The balanced accuracy and its posterior distribution.
\newblock In {\em 2010 20th International Conference on Pattern Recognition},
  pages 3121--3124, 2010.

\bibitem{bui2017hybrid}
Dieu~Tien Bui, Quang-Thanh Bui, Quoc-Phi Nguyen, Biswajeet Pradhan, Haleh
  Nampak, and Phan~Trong Trinh.
\newblock A hybrid artificial intelligence approach using gis-based
  neural-fuzzy inference system and particle swarm optimization for forest fire
  susceptibility modeling at a tropical area.
\newblock {\em Agricultural and forest meteorology}, 233:32--44, 2017.

\bibitem{EASTAUGH2014132}
C.S. Eastaugh and H.~Hasenauer.
\newblock Deriving forest fire ignition risk with biogeochemical process
  modelling.
\newblock {\em Environmental Modelling \& Software}, 55:132--142, 2014.

\bibitem{9554301}
Stella Girtsou, Alexis Apostolakis, Giorgos Giannopoulos, and Charalampos
  Kontoes.
\newblock A machine learning methodology for next day wildfire prediction.
\newblock In {\em 2021 IEEE International Geoscience and Remote Sensing
  Symposium IGARSS}, pages 8487--8490, 2021.

\bibitem{Huot2020segmentation}
Fantine Huot, R.~Lily Hu, Matthias Ihme, Qing Wang, John Burge, Tianjian Lu,
  Jason Hickey, Yi-Fan Chen, and John Anderson.
\newblock Deep learning models for predicting wildfires from historical
  remote-sensing data, 2020.

\bibitem{Jain2020}
Piyush Jain, Sean~C.P. Coogan, Sriram~Ganapathi Subramanian, Mark Crowley,
  Steve Taylor, and Mike~D. Flannigan.
\newblock A review of machine learning applications in wildfire science and
  management.
\newblock {\em Environmental Reviews}, 28(4):478--505, 2020.

\bibitem{prapas_deep_2021}
Ioannis Prapas, Spyros Kondylatos, Ioannis Papoutsis, Gustau Camps-Valls,
  Michele Ronco, Miguel-Ángel Fernández-Torres, Maria~Piles Guillem, and Nuno
  Carvalhais.
\newblock Deep {Learning} {Methods} for {Daily} {Wildfire} {Danger}
  {Forecasting}, November 2021.
\newblock Number: arXiv:2111.02736 arXiv:2111.02736 [cs].

\bibitem{ronneberger2015u}
Olaf Ronneberger, Philipp Fischer, and Thomas Brox.
\newblock U-net: Convolutional networks for biomedical image segmentation.
\newblock In {\em Medical Image Computing and Computer-Assisted
  Intervention--MICCAI 2015: 18th International Conference, Munich, Germany,
  October 5-9, 2015, Proceedings, Part III 18}, pages 234--241. Springer, 2015.

\bibitem{Santopaolo2021}
Alessandro Santopaolo, Syed~Saad Saif, Antonio Pietrabissa, and Alessandro
  Giuseppi.
\newblock Forest fire risk prediction from satellite data with convolutional
  neural networks.
\newblock In {\em 2021 29th Mediterranean Conference on Control and Automation
  (MED)}, pages 360--367, 2021.

\bibitem{zhang_forest_2019}
Guoli Zhang, Ming Wang, and Kai Liu.
\newblock Forest {Fire} {Susceptibility} {Modeling} {Using} a {Convolutional}
  {Neural} {Network} for {Yunnan} {Province} of {China}.
\newblock {\em International Journal of Disaster Risk Science}, 10(3):386--403,
  September 2019.

\end{thebibliography}

\end{document}